\documentclass[twoside,11pt]{article}

\usepackage{blindtext}
\usepackage{pmboxdraw}
\usepackage{jmlr2e}
\usepackage{lastpage}

\makeatletter
\def\@starteditor{}
\def\@endeditor{}
\def\@editor{}
\makeatother

\ShortHeadings{psifx feature extraction package}{Rochette, Vowels, Rochat}
\firstpageno{1}

\begin{document}

\title{psifx - Psychological and Social Interactions
Feature Extraction Package}
\author{
       \AND
       \name Guillaume Rochette\textsuperscript{*} \email guillaume.rochette@unil.ch \\
       \addr UNIL, Switzerland
       \AND
       \name Mathieu Rochat \email mathieu.rochat@unil.ch \\
       \addr UNIL, Switzerland
       \AND      
       \name Nizar Michaud \email nizar@ini.ethz.ch \\
       \addr Institute of Neuroinformatics, ETH Zurich, Switzerland
       \AND       
       \name Matthew J. Vowels\textsuperscript{*} \email matt@kivira.health \\
       \addr Kivira Health, New York, U.S.A. \\
       The Sense, CHUV, Switzerland \\
}
\makeatletter
\def\@makefntext#1{\noindent#1}
\makeatother
\footnotetext{*Equal contribution.}

\editor{}

\maketitle

\begin{abstract}
\href{https://psifx.github.io/psifx}{psifx} is a \emph{plug-and-play} multi-modal feature extraction toolkit, aiming to facilitate and democratize the use of state-of-the-art machine learning techniques for human sciences research. It is motivated by a need (a) to automate and standardize data annotation processes that typically require expensive, lengthy, and inconsistent human labor; (b) to develop and distribute open-source community-driven psychology research software; and (c) to enable large-scale access and ease of use for non-expert users. The framework contains an array of tools for tasks such as speaker diarization, closed-caption transcription and translation from audio; body, hand, and facial pose estimation and gaze tracking with multi-person tracking from video; and interactive textual feature extraction supported by large language models. The package has been designed with a modular and task-oriented approach, enabling the community to add or update new tools easily. This combination creates new opportunities for in-depth study of real-time behavioral phenomena in psychological and social science research.
\end{abstract}


\section{Introduction}
\label{sec:Introduction}

The study of human and social interactions requires access to meaningful and interpretable features representing these interactions. Currently, the standard approach involves human observational coding, which presents three major challenges: prohibitively high cost in time and training~\citep{Bulling2023}, lack of standardization across coders~\citep{Harris1982}, and limited scalability to large datasets. For example, manual transcription alone takes between five and ten times longer than the duration of the audio itself~\citep{Bazillon2008}. These problems hinder progress in behavioral research, particularly in mental health studies and intervention design.

To arrive at a set of variables representing complex video and audio data, one can deconstruct behavioral interactions into explainable features. Such features include \textit{non-verbal} behaviors (body and head pose, motion, gaze, facial expression), \textit{para-verbal} features (pitch, intonation), and \textit{verbal} features (transcribed language). These features represent directly observable events, in contrast with more elusive psychological constructs such as stress or emotion.

While state-of-the-art machine learning techniques are openly available for these tasks, including skeletal pose estimation~\citep{openpose}, facial analysis~\citep{openface2}, speech features~\citep{Eben2010opensmile}, and automatic transcription~\citep{Radford2023}, significant barriers to adoption exist in psychology and social science. These include technical hurdles in setup and deployment, the need for programming expertise, and concerns about data privacy when using third-party services.

To address these challenges, we present psifx (\underline{P}sychological and \underline{S}ocial \underline{I}nteractions \underline{F}eature e\underline{X}traction), an open-source project that provides:

\begin{itemize}
    \item An integrative approach to automate the extraction of objective non-verbal, para-verbal, and verbal features
    \item Efficient parallelism and hardware acceleration for processing large-scale datasets
    \item Simplified setup and usage through a homogeneous command-line interface
    \item A public, community-oriented repository with free-to-use Python package and Docker image
    \item Standardized task outputs using human-readable data formats
    \item Local data processing capabilities to protect sensitive information
\end{itemize}
Psifx is available here: \url{https://github.com/psifx/psifx}

\section{Features and Functionality}
\label{sec:Implementation}

psifx implements a modular architecture organized around three primary modalities: video, audio, and text. Each modality contains specialized modules that can be used independently or combined into processing pipelines. The system is designed to be extensible, allowing new modules to be added without affecting existing functionality.

\subsection{Installation and Usage}
\label{sec:Implementation:Installation}

One of the most important advantages that psifx has is in regard to the simplified installation process.
Despite the impressive performance of many open-source projects, the time and expertise required to install even just one such library can be significant.
The associated difficulties increase when one wishes to install multiple libraries simultaneously, creating issues with incompatibility between package dependencies. Indeed, the installation procedure for a (single) well-known computer vision package may require upwards of 20 commands involving the installation of a series of low-level dependencies.
In addition, psifx provides a Python package installable with pip, as well as a ready-to-use containerized images, in which the external libraries are guaranteed to be compatible with one another.
This provides a form of out-of-the-box practicality that many open-source projects simply do not have.

We provide a simple CLI for interfacing with the package. For example, to extract pose with mediapipe one can use the following command:

\begin{verbatim}
psifx video pose mediapipe multi-inference \
    --video input.mp4 \
    --poses output/poses.tar.gz \
    -- masks MaskD
\end{verbatim}

A Python interface is also available, although the principal use-case prioritizes users without programming experience.

\subsection{Video Processing}
\label{sec:Implementation:Video}

The video processing pipeline integrates multiple state-of-the-art tools for non-verbal feature extraction:

\subsubsection{Multi-Person Tracking}
Our current pipeline performs text-prompted multi-object video segmentation and tracking directly with SAM~3 \citep{carion2026sam3segmentconcepts}. To keep memory usage stable on long recordings, videos are processed in sequential frame chunks (default: 300 frames). Track identities are stitched across chunk boundaries by matching masks with an IoU criterion (default threshold: 0.3), so object IDs remain consistent over time. If a chunk causes CUDA out-of-memory, it is automatically split into smaller subchunks and retried. The tracker outputs one binary \texttt{.mp4} mask stream per global track, with empty-frame backfilling so all mask videos stay frame-aligned with the full source video. These per-track masks are then used for multi-person pose estimation by running MediaPipe separately on each masked (blackout) person stream, producing one pose archive per tracked individual.

\subsubsection{Human Pose Estimation}
We incorporate MediaPipe~\citep{Mediapipe} for real-time estimation of body, face, and hand configurations. This enables tracking of human movement and `body language' relevant to psychotherapy and behavioral analysis. It is  designed such that it can leverage the masks from the tracking algorithm for multi-person pose estimation.

\subsubsection{Facial Analysis}
OpenFace2.0~\citep{openface2} integration provides:
\begin{itemize}
    \item Gaze estimation
    \item Facial keypoints
    \item Facial action coding system (FACS) units
    \item Head pose estimation
\end{itemize}

It is designed such that it can leverage the masks from the tracking algorithm for multi-person pose estimation.

\subsection{Audio Processing}
\label{sec:Implementation:Audio}

The audio pipeline provides the following speech analysis capabilities:

\subsubsection{Speaker Diarization and Re-Identification}
We integrate pyannote~\citep{Bredin2023} for speaker diarization and implement a bespoke speaker re-identification system using ensemble embedding models. This allows robust mapping of speakers to their original audio channels in multi-microphone setups.

\subsubsection{Transcription and Analysis}
The system incorporates:
\begin{itemize}
    \item Whisper~\citep{Radford2023}, specifically the WhisperX implementation \citep{whisperx}, for multi-language transcription
    \item OpenSmile~\citep{Eben2010opensmile} for para-verbal feature extraction
    \item Enhanced transcription combining diarization and speaker identification
\end{itemize}

\subsection{Text Processing}
\label{sec:Implementation:Text}

Text processing capabilities leverage LangChain~\citep{Chase2022} to provide:
\begin{itemize}
    \item Flexible LLM integration (local or cloud-based)
    \item Generic instruction-based processing
    \item Interactive chat functionality
    \item Multiple model backends (Hugging Face, Ollama, OpenAI, Anthropic)
\end{itemize}

\section{Tests}
The package includes CI/CD workflows for automatic integration tests, and PyPI and Docker publishing.

\section{Data Quality and Hardware Guidelines}
\label{sec:Requirements}

For optimal performance, we recommend the use of synchronized multi-camera/microphone setups with single-person video frames for pose estimation. The microphones should be body-worn for clear audio separation between speakers. Adequate camera resolution and frame rate, and controlled/diffuse lighting and audio conditions are recommended.

The package supports both CPU-only and GPU-accelerated operation. CUDA-enabled hardware is recommended for large-scale data processing and local LLM hosting for the text-analysis tools.

\section{Development Roadmap}
\label{sec:Future}

Current development priorities include GUI design and implementation, integration of physiological sensing (e.g. respiration rate estimation from IR cameras), and multi-modal diarization approaches so that successful diarization can be achieved without multiple lavalier microphones.

\section{Conclusion}
\label{sec:Conclusion}

We have presented the open-source project psifx, an integrative package for multi-modal estimation of features relevant to psychological and social sciences.
The aim of psifx is to standardize and streamline the annotation processes of human interactions in order to increase reliability and reproducibility in this field of research.
But also to simplify and spread the uptake of state-of-the-art machine learning techniques in the community, by removing a number of technical barriers with respect to setup but also ease-of-use whilst retaining efficiency.
Moreover, the open-source and community-driven aspects will help to shaped and nurture organic growth and increase the longevity to the project.
We hope that psifx provides empirical researchers with usable, modern, open, and community driven extraction tools for non-verbal, para-verbal and verbal features.

\bibliography{sample}

@article{whisperx,
	author = {Bain, Max and Huh, Jaesung and Han, Tengda and Zisserman, Andrew},
	journal = {INTERSPEECH 2023},
	title = {WhisperX: Time-Accurate Speech Transcription of Long-Form Audio},
	year = {2023}}

@misc{ollama,
	author = {{Ollama}},
	title = {Ollama},
	url = {https://github.com/ollama/ollama},
	year = {2024}}

@misc{openai,
	author = {{Openai}},
	title = {Openai},
	url = {https://openai.com/},
	year = {2024}}

@misc{Chase2022,
	author = {Chase, H.},
	day = {17},
	month = {oct},
	title = {LangChain},
	url = {https://github.com/langchain-ai/langchain},
	year = {2022}}

@article{Bredin2023,
	author = {Bredin, H.},
	journal = {Proc. INTERSPEECH 2023},
	title = {pyannote.audio 2.1 speaker diarization pipeline: principle benchmark, and recipe},
	year = {2023}}

@webpage{Mediapipe,
	author = {{MediaPipe}},
	lastchecked = {10 May 2024},
	url = {https://developers.google.com/mediapipe/solutions/vision/pose_landmarker/},
	year = {2024}}

@article{Harris1982,
	author = {Harris, F.C. and Lahey, B.B.},
	doi = {https://doi.org/10.1016/0272-7358(82)90029-0},
	journal = {Clinical Psychology Review},
	number = {4},
	pages = {539-556},
	title = {Recording system bias in direct observational methodology: A review and critical analysis of factors causing inaccurate coding behavior.},
	volume = {2},
	year = {1982}}

@article{Bazillon2008,
	author = {Bazillon, T. and Est\`{eve}, Y. and Luzzati, D.},
	doi = {http://www.lrec-conf.org/proceedings/lrec2008/pdf/277_paper.pdf},
	journal = {Proceedings of the Sixth International Conference on Language Resources and Evaluation (LREC'08)},
	title = {Manual vs Assisted Transcription of Prepared and Spontaneous Speech},
	year = {2008}}

@article{Radford2023,
	author = {Radford, A. and Kim, J.W. and Xu, T. and Brockman, G. and Mcleavey, C. and Sutskever, I.},
	journal = {Proceedings of the 40th International Conference on Machine Learning, PMLR},
	pages = {28492-28518},
	title = {Robust Speech Recognition via Large-Scale Weak Supervision},
	volume = {202},
	year = {2023}}

@article{Bulling2023,
	author = {Bulling, L. and Heyman, R.E. and Bodenmann, G.},
	doi = {10.1037/fam0001036},
	journal = {Journal of Family Psychology},
	number = {1},
	pages = {1-9},
	title = {Bringing behavioral observation of couples into the 21st century},
	volume = {37},
	year = {2023}}

@article{Eben2010opensmile,
	author = {Eyben, F. and Wollmer, M. and Schuller, B.},
	doi = {10.1145/1873951.1874246},
	journal = {Proceedings 18th ACM international conference on multimedia},
	number = {1459-1642},
	title = {Opensmile: the munich versatile and fast open-source audio feature extractor},
	year = {2010}}

@article{openpose,
	author = {Cao, Z. and Hidalgo, G. and Simon T. and Wei, S.-E. and Sheikh, Y.},
	journal = {arXiv:1812.08008v1},
	title = {OpenPose: Realtime multi-person 2D pose estimation using part affinity fields},
	year = {2018}}

@article{gaze,
	author = {Kendon, A.},
	journal = {Acta Psychologica},
	pages = {1-47},
	title = {Some Functions of Gaze-Direction in Social Interaction},
	volume = {26},
	year = {1967}}

@article{openface2,
	author = {Baltrusaitis, T. and Zadeh, A. and Lim, Y. C. and Morency, L-P.},
	journal = {13th IEEE International Conference on Automatic Face and Gesture Recognition},
	title = {{OpenFace} 2.0: Facial Behavior Analysis Toolkit},
	year = {2018}}

@misc{carion2026sam3segmentconcepts,
      title={SAM 3: Segment Anything with Concepts}, 
      author={Nicolas Carion and Laura Gustafson and Yuan-Ting Hu and Shoubhik Debnath and Ronghang Hu and Didac Suris and Chaitanya Ryali and Kalyan Vasudev Alwala and Haitham Khedr and Andrew Huang and Jie Lei and Tengyu Ma and Baishan Guo and Arpit Kalla and Markus Marks and Joseph Greer and Meng Wang and Peize Sun and Roman Rädle and Triantafyllos Afouras and Effrosyni Mavroudi and Katherine Xu and Tsung-Han Wu and Yu Zhou and Liliane Momeni and Rishi Hazra and Shuangrui Ding and Sagar Vaze and Francois Porcher and Feng Li and Siyuan Li and Aishwarya Kamath and Ho Kei Cheng and Piotr Dollár and Nikhila Ravi and Kate Saenko and Pengchuan Zhang and Christoph Feichtenhofer},
      year={2026},
      eprint={2511.16719},
      archivePrefix={arXiv},
      primaryClass={cs.CV},
      url={https://arxiv.org/abs/2511.16719}, 
}

\end{document}